\documentclass[letterpaper, 10 pt, journal, twoside]{IEEEtran}

\IEEEoverridecommandlockouts %

\usepackage{graphics} %
\usepackage{amsmath} %
\usepackage{amssymb} %
\usepackage{cite}
\usepackage[usenames,dvipsnames]{xcolor}
\usepackage{graphicx}
\usepackage{booktabs}
\usepackage{tabularx}
\usepackage{multirow}
\usepackage{amsfonts}
\usepackage{mathtools}
\usepackage[binary-units=true]{siunitx}
\usepackage[hidelinks]{hyperref}
\usepackage[T1]{fontenc} %
\usepackage{mathtools, cuted} %
\usepackage[ruled,vlined]{algorithm2e} 
\usepackage{balance} %
\usepackage{enumerate}  
\usepackage{array}
\usepackage[export]{adjustbox} %

\newcolumntype{x}[1]{%
>{\raggedleft\hspace{0pt}}p{#1}}%

\usepackage{subcaption}
\DeclareCaptionLabelSeparator{periodspace}{.\quad}
\newcommand{\ra}[1]{\renewcommand{\arraystretch}{#1}}

\usepackage{amsthm}

\newtheorem{prop}{Proposition}

\theoremstyle{definition}

\usepackage{letltxmacro}
\LetLtxMacro\orgvdots\vdots
\LetLtxMacro\orgddots\ddots

\makeatletter
\DeclareRobustCommand\vdots{%
	\mathpalette\@vdots{}%
}
\newcommand*{\@vdots}[2]{%
	\sbox0{$#1\cdotp\cdotp\cdotp\m@th$}%
	\sbox2{$#1.\m@th$}%
	\vbox{%
		\dimen@=\wd0 %
		\advance\dimen@ -3\ht2 %
		\kern.5\dimen@
		\dimen@=\wd2 %
		\advance\dimen@ -\ht2 %
		\dimen2=\wd0 %
		\advance\dimen2 -\dimen@
		\vbox to \dimen2{%
			\offinterlineskip
			\copy2 \vfill\copy2 \vfill\copy2 %
		}%
	}%
}
\DeclareRobustCommand\ddots{%
	\mathinner{%
		\mathpalette\@ddots{}%
		\mkern\thinmuskip
	}%
}
\newcommand*{\@ddots}[2]{%
	\sbox0{$#1\cdotp\cdotp\cdotp\m@th$}%
	\sbox2{$#1.\m@th$}%
	\vbox{%
		\dimen@=\wd0 %
		\advance\dimen@ -3\ht2 %
		\kern.5\dimen@
		\dimen@=\wd2 %
		\advance\dimen@ -\ht2 %
		\dimen2=\wd0 %
		\advance\dimen2 -\dimen@
		\vbox to \dimen2{%
			\offinterlineskip
			\hbox{$#1\mathpunct{.}\m@th$}%
			\vfill
			\hbox{$#1\mathpunct{\kern\wd2}\mathpunct{.}\m@th$}%
			\vfill
			\hbox{$#1\mathpunct{\kern\wd2}\mathpunct{\kern\wd2}\mathpunct{.}\m@th$}%
		}%
	}%
}
\makeatother

\def\br{\mathbb R}

\def\bs{\mathbb S}

\def\ba{\mathbb A}
\def\bs{\mathbb S}

\def\sg{\mathcal{G}}
\def\sa{\mathcal{A}}

\def\sl{\mathcal{L}}

\def\tr{\mathrm{tr}}

\newcommand\eqdef{\mathrel{\overset{\makebox[0pt]{\mbox{\normalfont\tiny def}}}{=}}}

\renewcommand{\r}[1]{{\color{red}{#1}}}

\graphicspath{{Figs/}}

\newcommand{\edittwo}[1]{\textcolor{black}{#1}}
\newcommand{\edit}[1]{\textcolor{black}{#1}}

\title{%
	 A Distributed Pipeline for \\ Scalable, Deconflicted Formation Flying
	}

\author{Parker C. Lusk, Xiaoyi Cai, Samir Wadhwania, Aleix Paris, Kaveh Fathian, Jonathan P. How%
    \thanks{Manuscript received: February 24, 2020; Revised: May 20, 2020; Accepted: June 1, 2020}%
	\thanks{
	    This paper was recommended for publication by N.Y.\ Chong upon evaluation of the Associate Editor and Reviewers' comments. %
	    Research supported in part by NASA Convergent Aeronautics Solutions project Design Environment for Novel Vertical Lift Vehicles (DELIVER),
		ARL DCIST under Cooperative Agreement Number W911NF-17-2-0181,
		and Boeing Research \& Technology.
		Computation support provided by Amazon Web Services.}
	\thanks{P.\ C.\ Lusk, X.\ Cai, S.\ Wadhwania, A.\ Paris,  K.\ Fathian and J.\ P.\ How are with the Department of Aeronautics and Astronautics, Massachusetts Institute of Technology.
	    {\{plusk, xyc, samirw, aleix, kavehf, jhow\}@mit.edu.}}
	\thanks{Digital Object Identifier (DOI): see top of this page.} %
}%

\begin{document}

\maketitle

\begin{abstract}
Reliance on external localization infrastructure and centralized coordination are main limiting factors for formation flying of vehicles in large numbers and in unprepared environments.
While solutions using onboard localization address the dependency on external infrastructure, the associated coordination strategies typically lack collision avoidance and scalability.
To address these shortcomings, we present a unified pipeline with onboard localization and a distributed, collision-free formation control strategy that scales to a large number of vehicles.
Since distributed collision avoidance strategies are known to result in gridlock, we also present a \edit{distributed} task assignment solution to deconflict vehicles.
We experimentally validate our pipeline in simulation and hardware.
The results show that our approach for solving the optimization problem associated with formation control gives solutions within seconds in cases where general purpose solvers fail due to high complexity.
In addition, our lightweight assignment strategy leads to successful and quicker formation convergence in
\SIrange[range-units=single,range-phrase=--]{96}{100}{\percent} of all trials,
whereas indefinite gridlocks occur without it for \SIrange[range-units=single,range-phrase=--]{33}{50}{\percent} of trials.
By enabling large-scale, deconflicted coordination, this pipeline should help pave the way for anytime, anywhere deployment of aerial swarms.
\end{abstract}

\vspace{-0.8em}

 \section*{Supplementary Material}

Video and open-source implementation available at \href{https://github.com/mit-acl/aclswarm}{\texttt{https://github.com/mit-acl/aclswarm}}.

\section{Introduction}\label{sec:intro}

\IEEEPARstart{T}{wo} %
main challenges in the deployment of large-scale swarms are the localization and coordination of vehicles.
Localization methods that rely on external infrastructure 
(e.g., GPS) 
are prone to systematic errors (e.g., multipath effect)
and may not always be available.
Coordination strategies that are centralized can deconflict motion plans to prevent collisions and gridlock, but introduce a single point of failure and are difficult to scale in swarm size due to communication bandwidth limitations.

This paper presents a unified formation flying pipeline for unmanned aerial vehicles (UAVs).
Our pipeline uses \textit{onboard} sensors for localization, which eliminate the need for external positioning systems, and \textit{distributed} techniques for coordination, which enable each vehicle to make decisions independently while communicating their state to a subset of the team.
For \textit{localization}, we use an off-the-shelf commercial visual inertial odometry (VIO) package \cite{VIO}
that fuses inertial measurement unit (IMU) and downward-facing monocular camera measurements to estimate changes in the vehicle pose.
\edit{For \textit{coordination}, we present distributed formation control and task assignment strategies that run onboard the vehicles, do not rely on a common reference frame, and use vehicle-to-vehicle communication.} 
Key features of our formation control strategy include scalability to a large number of vehicles and robustness to disturbances.
The latter is crucial for reaching the desired formations with sensing imperfections.
Our task assignment strategy uses an auction-based algorithm to guarantee conflict-free assignments.
This algorithm can deconflict vehicle gridlocks resulting from distributed collision avoidance (type 3 deadlock~\cite{Wang2017}) and is well-suited for vehicles with limited computational capability and low-bandwidth communication.

\begin{figure}[t!]
	\begin{center}
		\includegraphics[trim =0mm 10mm 0mm 0mm, clip, width=\columnwidth]{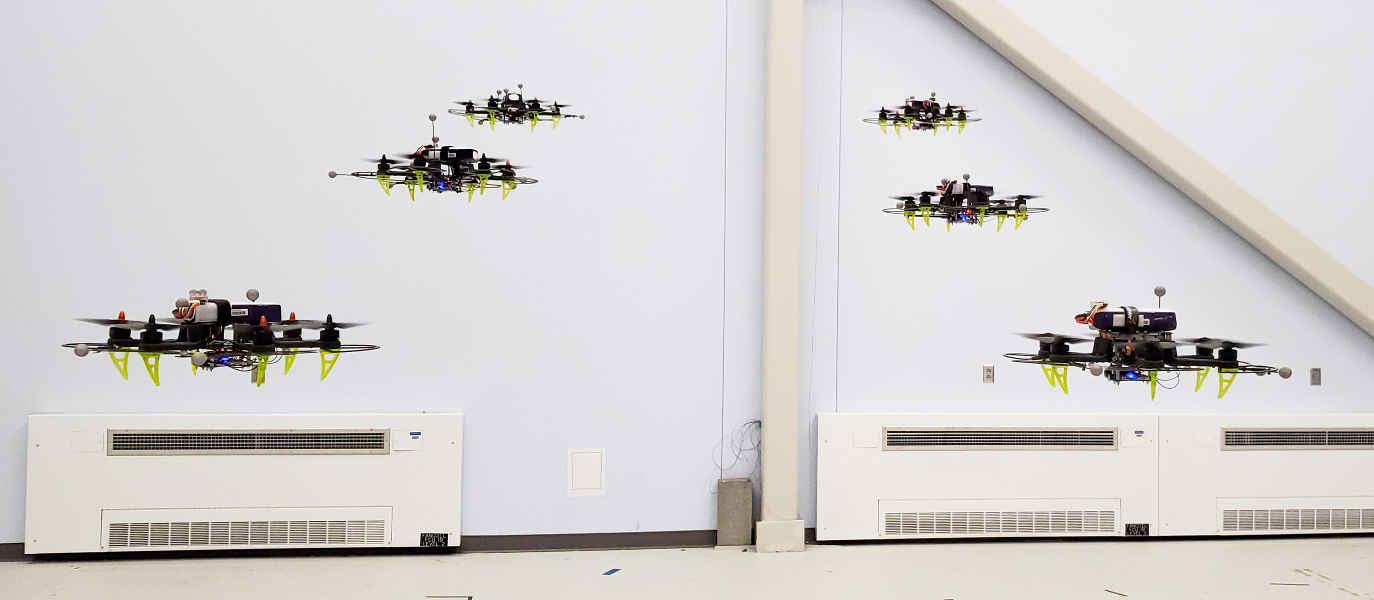}	
		\caption{
		Six multirotors in a slanted plane formation.
		Vehicles communicate with each other, make distributed decisions onboard, and use VIO for localization.}
		\label{fig:slantedplane}
	\end{center}
\end{figure}

\subsection{Contributions}

This research extends our previous work on UAV formations~\cite{Fathian2019} and presents a unified pipeline consisting of \textit{onboard localization} and \textit{distributed coordination}.
The three main contributions of this work are:
\begin{enumerate}
    \item \edit{scalable formulation of control design suitable for
    onboard sensing without a common reference frame;}
    \item algorithms for deconfliction via \edit{distributed} task assignment of vehicles to desired formation points;    
    \item simulation- and hardware-ready open-source pipeline.
\end{enumerate}
\edit{Our pipeline is tested in hardware with six multirotors (see Fig.~\ref{fig:slantedplane}), and 
to our knowledge is the first demonstration of formation flying that does not rely on external sensing, fiducial markers for localization, a common reference frame, or a centralized base station for coordination.}
The only requirements for the presented pipeline are that the vehicles can communicate, can find the transformation between their VIO start frames, and the environment is sufficiently textured---a standard assumption for VIO systems.
As such, this framework paves the way for future, real-world deployments of aerial vehicle swarms in large numbers and without requiring external localization infrastructure.

\begin{figure} [t!]
\centering
	\begin{subfigure}[b]{0.32\columnwidth}
	    \includegraphics[width=0.8\textwidth,left]{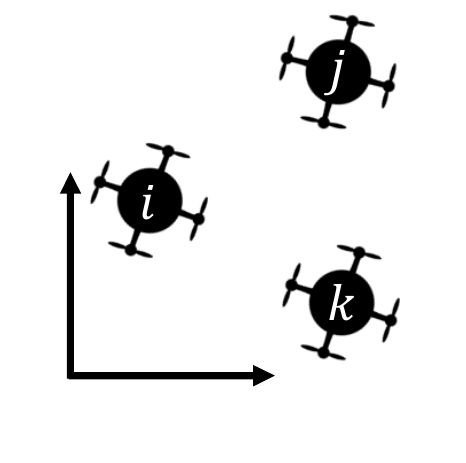}
	    \caption{\scriptsize full alignment}
	    \label{fig:frame-a}
	\end{subfigure}
	\begin{subfigure}[b]{0.32\columnwidth}
	    \includegraphics[width=0.8\textwidth,center]{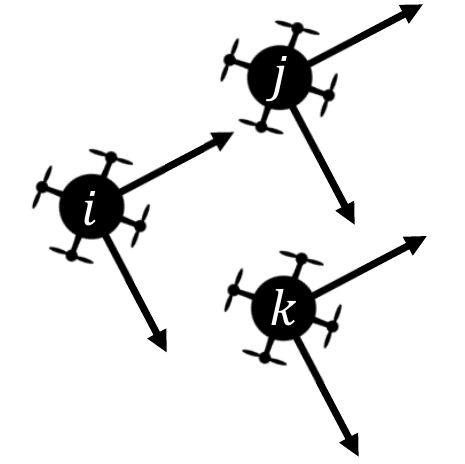}
	    \caption{\scriptsize orientation alignment}
	    \label{fig:frame-b}
	\end{subfigure}
	\begin{subfigure}[b]{0.32\columnwidth}
	    \includegraphics[width=0.8\textwidth,right]{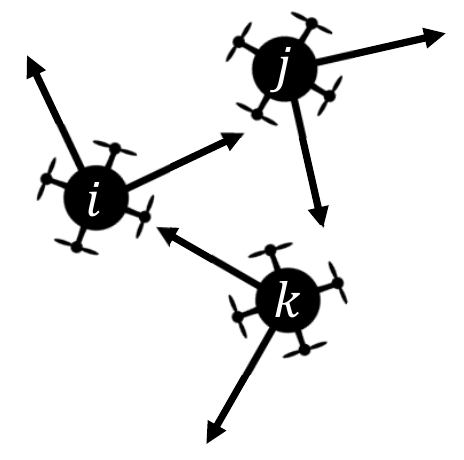}
	    \caption{\scriptsize no alignment}
        \label{fig:frame-c}
	\end{subfigure}
\caption{\edit{Required alignment of UAV frames in existing swarm strategies: (a) the most restrictive case requiring a common reference frame, i.e., orientation and origin of the frames must be aligned; (b) only the orientation of the frames must be aligned; (c) no alignment restrictions (this work).}}
	\label{fig:Frames}
\end{figure}

\subsection{Related Work}

Existing aerial swarms can be grouped based on the coordination (centralized vs.\ distributed) and localization (external vs.\ onboard) methods used. 
\edit{It is further crucial to distinguish these methods based on the level of alignment required for the vehicle coordinate frames; see Fig.~\ref{fig:Frames}.} 
 
\edit{
Works with \textit{centralized} coordination and \textit{external} localization include~\cite{Preiss2017, Honig2018, Du2019}, which are based on lightweight UAVs with limited onboard computational capability and therefore rely on an external motion capture system and a base station.
Works with \textit{distributed} coordination and \textit{external} localization include \cite{wilson2020robotarium}, \cite{enright2004spheres}, where robots execute distributed controls  based on external localization by motion capture and ultrasonic beacons, respectively.
Works with \textit{centralized} coordination and \textit{onboard} localization include~\cite{Forster2013}, \cite{Loianno2016}, which use a ground station for task assignment among vehicles.
In \cite{Weinstein2018}, formation flying based on VIO is demonstrated, where motion planning and assignment are run on a base station to ensure collision-free trajectories.
The coordination strategies used in aforementioned works require a \textit{common reference frame} (Fig.~\ref{fig:frame-a}).
}

\edit{
Despite the large body of work on formation control~\cite{Oh2015}, and the variety of onboard sensing solutions for localization (e.g., VIO~\cite{Delmerico2018}), few frameworks demonstrated formation flying with \textit{distributed} coordination and \textit{onboard} localization.
A key reason is reliance of many distributed control and assignment algorithms on aligned frames (Fig.~\ref{fig:frame-a}, \ref{fig:frame-b}), which require computation-expensive and/or communication-intensive synchronization/consensus steps for frame alignment.
Equally important, dependence on alignment in existing methods \cite{Wang2017,Turpin2014, van2011reciprocal, morgan2016swarm} diminishes robustness to inherent noise and unobservable errors that cannot be corrected (e.g., disparities between the actual and estimated body frame \textit{orientation} caused by VIO drift).
Leveraging coordination methods that are \textit{robust to misaligned frames} is hence crucial and a focus of this work. 
}

\edit{
Examples of other pipelines with distributed coordination and onboard localization include \cite{Montijano2016,Tron2016}.
Both works demonstrated formation flying on three UAVs, required information from an external motion capture system due to hardware limitations, did not incorporate collision avoidance, and required frame alignment.
}
\edittwo{Note that while~\cite{Montijano2016,Tron2016} can achieve formations with arbitrary headings as illustrated in Fig.~\ref{fig:frame-c}, knowledge of relative orientations is still required; therefore, they belong to the category of Fig.~\ref{fig:frame-b}.}

\if 0

\r{
decentralized coordination setting combined with VIO:
D-CAPT [26]~\cite{}:
ORCA ~\cite{}: 
CBF [2]~\cite{} :
[A]
}

\r{Robusteness in coordination,  with compounded noise/latency, which would eventually break (b).\\

some existing algorithm might as well
work in a similar fully decentralized setting, when combined with VIO
as proposed here. For example, D-CAPT [26], ORCA, CBF [2] might also be
useful for such a task and are computationally even more efficient than
the proposed approach. \\

R2:  onboard sensing for localization ->
 Finally, the related work section only
focuses on this aspect of the pipeline, discussing how many formation papers include
onboard localization but barely sells the advantages of the coordination module (the actual
proposal of the paper) against other competitors such as [26] or [A] or to mention similar
coordination pipelines. \\

Given a solution to this problem, the controller in Section III seems unnecessary, each drone
has a target position and can use a local controller with collision avoidance that drives it to
that position. Note that such controllers exists in the literature (e.g., RVO in any of its
multi-agent variantes), they are distributed in nature and only require local sensing.

}

\fi

\section{System Overview}\label{sec:overview}

\begin{figure}[t!]
	\begin{center}
		\includegraphics[trim =0mm 0mm 0mm 0mm, clip, width=\columnwidth]{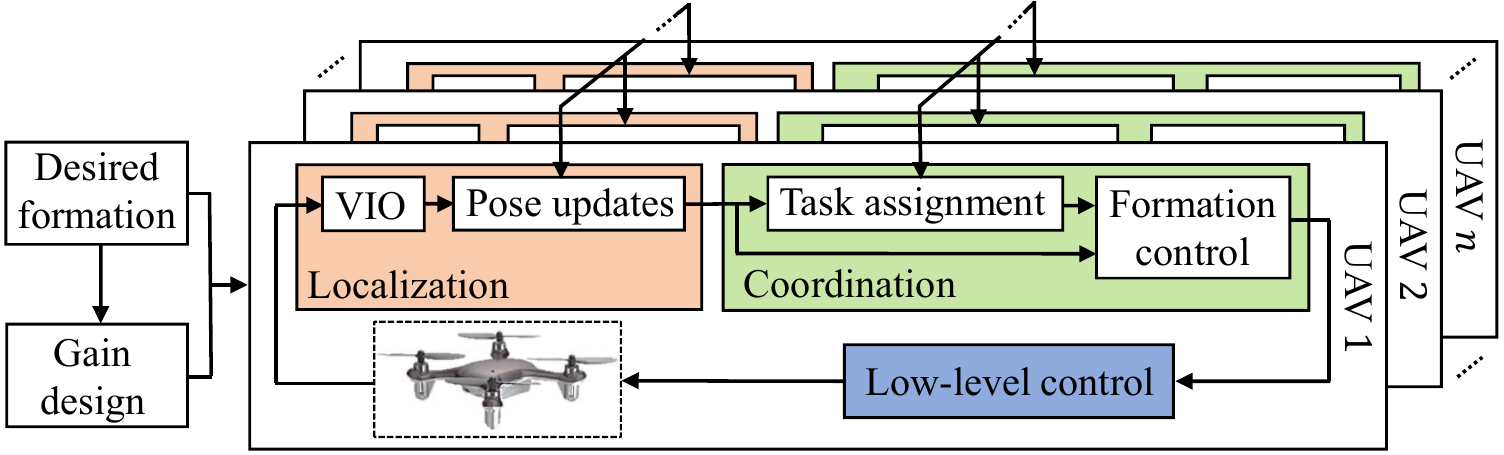}	
		\caption{\edit{Modules of our formation flying pipeline. The desired formation is used to design required gains for formation control. The localization framework provides self and relative pose measurements. The coordination framework assigns each UAV to a formation point and plans its motion to attain formation flying.}}
		\label{fig:System}
	\end{center}
\end{figure}

A schematic representation of our pipeline is depicted in Fig.~\ref{fig:System} for a swarm of $n$ multirotor UAVs.
The key components of this pipeline include modules for localization and coordination of the vehicles, which require exchanging information between a subset of UAV peers referred to as \textit{neighbors}.
\edit{The main goal of the pipeline is formation flying.
We assume a desired formation shape is specified by an operator.
This desired formation is used to design the required gains for formation control and both are given as input to the vehicles.
}

\edit{
With onboard localization,
the pose of a UAV with respect to its own start frame, which is fixed at its initial pose, is estimated using VIO.}
These self pose estimates provide feedback to the low-level controller, which stabilizes the UAV and tracks a reference velocity specified by the high-level formation control strategy.
Through inter-vehicle pose updates, a UAV acquires relative pose estimates of its neighbors by transforming their poses into its own start frame.
This process requires knowledge of the transformations that relate the UAVs' start frames.
Several methods can be used to obtain these transformations. 
For instance, if two vehicles have a common field of view, once the correspondence among the 3D landmarks reconstructed by VIO is determined, the relative pose between the UAVs' start frames can be found using Arun's method~\cite{arun1987least}.
For simplicity, the transformations are obtained in our experiments by initializing the UAVs at pre-specified locations.
\edit{
Note that the UAVs do \textit{not} require the transformations to \textit{non-neighboring} vehicles.
}

\edit{The coordination framework handles formation flying of the UAV swarm.
This framework consists of the task assignment and formation control modules, as depicted in Fig.~\ref{fig:System}}.
Formation control is concerned with finding collision-free trajectories that bring the UAVs to a desired formation.
A \textit{desired formation} is defined by a graph $\sg$ with vertices located at 3D points $p_1, \dots, p_n$ and edges connecting the vertices to indicate neighbors.
\edit{Graph $\sg$ is used in designing the formation control and is also 
broadcasted to the UAVs from the base station.}
The vehicles aim to achieve the overall geometric shape specified by the points $p_i$ (rather than the exact location and orientation of this point configuration in the space).
Throughout this paper, we assume that $\sg$ is undirected, connected, and \textit{universally rigid} \cite{Gortler2014}.
Informally, rigidity implies that $\sg$ cannot be deformed without violating the desired distances between formation points.

The goal of task assignment is to uniquely allocate each UAV to a point in the desired formation.
Each UAV $i$ is assigned to a formation point $p_j$ using a one-to-one assignment map $\sigma$ with $\sigma(i) = j$ (see Section~\ref{sec:assign}). 
The set of neighbors of UAV $i$, denoted by $\mathcal{N}_i$, is defined as the set of UAVs $j$ such that $p_{\sigma(j)}$ is connected to $p_{\sigma(i)}$ by an edge in $\sg$.
UAV $i$ and its neighbors communicate to attain relative pose measurements using the localization framework.
\edittwo{We emphasize that $\mathcal{N}_i$ is defined according to the current assignment map $\sigma$ and formation graph $\sg$, and that these neighbors are used for both formation control and communication.
As tasks are reassigned, the set of neighbors, and therefore communication links, may change.}

\section{Distributed Formation Control} \label{sec:motionplanning}

\edit{
To achieve a given desired formation, we require a distributed strategy in which the UAVs execute their motions independently and are robust to misalignments of UAV frames.}
To make the paper self-contained, we first review a candidate strategy and then present a solution to address the scalability issue that arises for large-scale formations.

\subsection{Overview of Formation Control}

The framework for achieving a formation using only relative and local position measurements is based on \cite{Lin2016, Lin2016a} and our previous work \cite{Fathian2019, Fathian2018b}. 
For each UAV, the key steps in this strategy can be summarized as follows.
\begin{enumerate}
\item
The UAV calculates the position vectors from itself to each of its neighbors in its own body frame.
\item Each vector is scaled and rotated about the $z$-axis of the UAV's body frame. The amount of scaling and rotation is pre-specified and depends on the desired formation. 
\item These scaled and rotated vectors are then summed to obtain a resultant velocity vector command.
\end{enumerate}

\edit{We emphasize that the above strategy does \textit{not} rely on a common reference frame  (Fig.~\ref{fig:frame-a}) and the scaling and rotation are performed in each respective UAV body frame.} To formulate and analyze this framework mathematically, however, we consider a common reference frame, in which we express the position of UAV $i$ by $q_i \in \br^3$ and the  vector connecting UAV $i$ to its neighbor $j \in \mathcal{N}_i$  by $q_j - q_i$. The scaling and rotation of this vector is expressed as $A_{ij} (q_j - q_i)$, where $A_{ij} \in \ba(3)$ is called a \textit{gain matrix} and belongs to the set of scaled rotation matrices along the $z$-axis denoted by 
\begin{equation} \label{eq:Aset}
\ba(3) \eqdef \left\{ \begin{bsmallmatrix}
a & -b & \; 0 \\
b & ~a & \; 0 \\
0 & ~0 & \; c \\
\end{bsmallmatrix} \,:\, a,b,c \in \br  \right\}.
\end{equation}
Consequently, the motion of UAV $i$ can be expressed as 
\begin{equation} \label{eq:control}
\dot{q}_i = \sum_{j \in \mathcal{N}_i}{A_{ij} \, (q_j - q_i)},
\end{equation}
where $\dot{q}_i$ is the velocity vector that encapsulates the desired speed and direction of motion for the vehicle.
\edittwo{While we consider single-integrator dynamics for simplicity of motion planning, higher-order dynamics can be utilized~\cite{Fathian2018b}.}

In \eqref{eq:control}, we assume that the $z$-axes of UAVs' body frames (and the reference frame used for the analysis) are aligned. In practice, the direction of gravity can be used to align these axes, and, as we will discuss in Section~\ref{sec:robustness}, small misalignments caused by measurement errors or acceleration effects do not affect the convergence. 
\edit{
Note that we do not require that the $x$-$y$ axes be aligned; the UAVs can have arbitrary yaw orientations (Fig.~\ref{fig:frame-c}). This point distinguishes \eqref{eq:control} from the consensus-based \cite{Ren2007}, bearing-based \cite{Zhao2019}, or similar distributed control \cite{Montijano2014}, in which convergence guarantees rely on orientation alignment or consensus of the UAV body frames (Fig.~\ref{fig:frame-b}). }

To analyze the trajectory of the swarm, we define  
\begin{equation} \label{eq:qA}
q \eqdef \begin{bsmallmatrix}
q_{1} \\
q_{2} \\
\vdots \\
q_{n}   
\end{bsmallmatrix}, \quad 
A \eqdef \begin{bsmallmatrix}
-\sum_{j} A_{1j} & A_{12} & \cdots & A_{1n} \\
A_{21} & -\sum_{j} A_{2j} & \cdots & A_{2n} \\
\vdots &                          & \ddots & \vdots \\
A_{n1} &       A_{n2}             & \cdots &  -\sum_{j} A_{nj}   
\end{bsmallmatrix},
\end{equation}
where $q$ is the aggregate vector of UAV positions and $A$ consists of gain matrices. Here, if UAVs $i$ and $j$ are not neighbors, the corresponding $A_{ij}$ is defined as a zero matrix.
Based on \eqref{eq:control} and by using the notation in \eqref{eq:qA}, swarm motion can be expressed by $\dot{q} = A\, q$, which determines the trajectories that the UAVs traverse.

Given a desired formation expressed via the set of points ${p_i = [x_i,\, y_i, \, z_i]^\top} \in \br^3$, we define
\begin{equation} \label{eq:N}
N \eqdef \begin{bsmallmatrix}
p^x_1 & p^y_1 & p^z_1 &  e^x & e^y & e^z \\ 
p^x_2 & p^y_2 & p^z_2 &  e^x & e^y & e^z \\
\vdots &  \vdots & \vdots & \vdots & \vdots & \vdots \\ 
p^x_n & p^y_n & p^z_n &  e^x & e^y & e_z 
\end{bsmallmatrix} \in \mathbb{R}^{3n\times 6},
\end{equation}
where ${p^x_i \eqdef [x_i,\, y_i,\, 0]^\top}$, ${p^y_i \eqdef [-y_i,\, x_i,\, 0]^\top}$, ${p^z_i \eqdef [0,\, 0,\, z_i]^\top}$, ${e^x \eqdef [1,\,0,\,0]^\top}$, ${e^y \eqdef [0,\,1,\,0]^\top}$, and ${e^z \eqdef [0,\,0,\,1]^\top}$. 
Convergence to the desired shape is guaranteed if $A\, N = 0$ (i.e., columns of $N$ are null vectors of $A$) and if all remaining eigenvalues of $A$ not associated with $N$ are strictly negative.
For such an $A$ to exist, each UAV should have a sufficient number of neighbors. Specifically, if vertices and edges represent the UAVs and their neighboring relations in the formation graph $\sg$, $A$ exists if $\sg$ is universally rigid (see~\cite[Thm. 3.2]{Lin2016a}).

\edit{Our approach consists of \textit{design} and \textit{execution} phases.
In the \textit{design} phase, a gain matrix $A$ as in~\eqref{eq:qA} is computed offline based on the specification of a formation \edittwo{graph $\sg$} and serves as an input to the distributed algorithms that run onboard the UAVs.
The \textit{execution} phase is entirely distributed, where the UAVs plan their trajectories independently using relative translation measurements to their neighboring UAVs.}

\subsection{Scalable Gain Design}

Given a desired formation, the gain matrix $A$ that meets the aforementioned constraints can be computed from
\begin{equation} \label{eq:OptimCVX}
\begin{aligned}
& \underset{A \in \bs^{-}_{3n}}{\text{minimize}}
& & \lambda_{\text{max}} \left( Q^\top A \, Q \right) & \\
& \text{subject to}
& & A\, N = 0 & \\
&&& A_{ij} \in \mathbb{A}(3)  & \forall_{i,j} \\
&&& A_{ij} = 0 \;   & \forall_{i} \ \forall_{j \notin \mathcal{N}_i} \\
&&& \tr(A) = \text{constant}
\end{aligned}
\end{equation}
where $\lambda_{\text{max}}$ denotes the largest eigenvalue of a matrix, $Q \in \mathbb{R}^{3n\times (3n-6)}$ is the orthogonal complement of $N$ (i.e., $N^\top Q = 0$), which is found from the singular value decomposition of $N$, and $\bs^-_{3n}$ is the space of symmetric negative semidefinite matrices of dimension $3n$.
The objective of \eqref{eq:OptimCVX} is to make the nonzero eigenvalues of $A$ as negative as possible ($Q^\top A \, Q$ is the restriction of $A$ on the subspace $Q$ and eliminates the zero eigenvalues associated with $N$).
By doing so, stability and robustness of the formation to noise, measurement errors, and disturbances is increased.  
We note that the last constraint in \eqref{eq:OptimCVX} sets the trace of $A$ to a constant value to ensure that the problem is bounded (without this constraint, if $A\in\bs^-_{3n}$ is a solution, so is $c\, A$ for any $c>0$ with a better objective value).
The universal rigidity assumption on the formation graph \cite[Thm. 3.2]{Lin2016a} is sufficient to ensure that \eqref{eq:OptimCVX} is feasible and that all remaining eigenvalues of $A$ not associated with $N$ are strictly negative.

The formulation \eqref{eq:OptimCVX} was presented in our earlier work~\cite{Fathian2019} and can be solved relatively quickly using existing SDP solvers for small number of vehicles. However, for large-scale problems (e.g., more than $50$ UAVs) it becomes challenging, or even impossible, to solve. 
We address this issue by exploiting the problem structure to derive a solution based on the alternating direction method of multipliers (ADMM).

We observe from \eqref{eq:Aset} that $A_{ij}\in\ba(3)$ has a block diagonal structure, which can be expressed by ${A_{ij} = \mathrm{blkdiag}(D_{ij}, c_{ij})}$, where the $2\times 2$ matrix $D_{ij}$ consists of the first two rows and columns, and the scalar $c_{ij}$ is the entry in the last row and column of $A_{ij}$.
Due to this structure, we conclude from \eqref{eq:control} that vehicle trajectories along the $x$-$y$  and $z$ components are decoupled and depend only on $D_{ij}$ and $c_{ij}$, respectively.
This observation allows us to split \eqref{eq:OptimCVX} into two subproblems with lower dimensions, leading to reduced computational effort.
By defining
\begin{equation} \label{eq:} 
B \eqdef \begin{bsmallmatrix}
-\sum_{j} c_{1j} & c_{12} & \cdots & c_{1n} \\
c_{21} & -\sum_{j} c_{2j} & \cdots & c_{2n} \\
\vdots &                      & \ddots & \vdots \\
c_{n1} &     c_{n2}       & \cdots &  -\sum_{j} c_{nj}   
\end{bsmallmatrix}, ~~
M \eqdef \begin{bsmallmatrix}
z_1 & 1 \\ 
z_2 & 1 \\
\vdots &  \vdots \\ 
z_n & 1
\end{bsmallmatrix},
\end{equation}
which correspond to the $z$ components of $A$ in \eqref{eq:qA} and $N$ in \eqref{eq:N}, the problem of finding $c_{ij}$ is formulated as
\begin{equation} \label{eq:OptimZ}
\begin{aligned}
& \underset{B \in \bs^{-}_{n} }{\text{minimize}}
& & \lambda_{\text{max}} \left(  R^\top B \, R \right) & \\
& \text{subject to}
& & B \, M = 0 & \\
&&& c_{ij} = 0 \;   & \forall_{i} \ \forall_{j \notin \mathcal{N}_i} \\
&&& \tr(B) = \text{constant}
\end{aligned}
\end{equation}
where $R \in \br^{n \times (n-2)}$ is the orthogonal complement of $M \in \br^{n\times 2}$. 
The optimization problem for finding $D_{ij}$ is formulated similarly to \eqref{eq:OptimZ}, with an additional constraint that the diagonal entries of $D_{ij}$ must be equal and the off-diagonal entries must have the same absolute value with different signs. With this point in mind, we henceforth focus our attention on \eqref{eq:OptimZ}. 
The following proposition brings \eqref{eq:OptimZ} into the standard form suitable for applying ADMM.

\begin{prop} \label{prop:SDP}
Problem \eqref{eq:OptimZ} can be formulated as
\begin{equation} \label{eq:OptimX}
\begin{aligned}
& \underset{X \in \bs^+_{2n-4} }{\text{minimize}}
& & \left< C,\, X \right> & \\
& \text{subject to}
& & \sa(X) = b  
\end{aligned}
\end{equation}
where $\left< C,\, X \right> \eqdef \tr(C^\top X)$,  
$C \eqdef \begin{bsmallmatrix}
I & 0 \\
0 & 0
\end{bsmallmatrix}$ with $I$ as the identity matrix of size $n-2$, and $b \in \br^m$. The operator $\sa(X)$ represents a set of linear constraints on $X$ and enforces it to have the block diagonal structure $X \eqdef \begin{bsmallmatrix}
\gamma\, I && I \\
I && Z 
\end{bsmallmatrix}$, where $\gamma \geq 0$ and $Z \in \bs^+_{n-2}$. The solution of \eqref{eq:OptimZ} is obtained from $X$ as $B = - M \, Z \, M^\top$.

\end{prop}

\edit{Proposition~\ref{prop:SDP} is proved in the appendix of~\cite{lusk2020distributed}.} 
We now leverage the ADMM technique in \cite{Wen2010} to solve  \eqref{eq:OptimX}.
From~\cite{Wen2010}, the augmented Lagrangian associated with \eqref{eq:OptimX} is 
\begin{equation} \label{eq:Lagrangian}
\mathcal{L} \eqdef - \langle y, b \rangle + \langle \sa^*(y) + S - C,\, X \rangle + \frac{1}{2\mu} \| \sa^*(y) + S - C \|,
\end{equation}
where $S \in \bs^+_{2n-4}$ and $y$ are dual variables associated with constraints $X \in \bs^+_{2n-4}$ and $\sa(X) = b$, respectively, $\sa^*$ is the adjoint of $\sa$, and $\mu > 0$ is a penalty parameter that balances the standard Lagrangian and the augmented term. 
ADMM then proceeds by alternatively optimizing each primal and dual variable with others fixed, which results in a closed-form solution for each subproblem. Denoting by the superscript $k$ the iteration number, the ADMM iterative update procedure is given as
\begin{equation} \label{eq:ADMM}
\begin{aligned}
y^{k+1} &= (\sa \sa^*)^{-1} \left( \sa (C - S^k - \mu X^k) + \mu\, b \right), \\
W^{k+1} &=  C - \sa^* (y^{k+1}) - \mu \, X^k, \\
S^{k+1} &= \mathcal{P}_{\mathrm{psd}} \left(W^{k+1} \right), \\
X^{k+1} 
&= \frac{1}{\mu} \left( S^{k+1} - W^{k+1} \right).
\end{aligned}
\end{equation}
In~\eqref{eq:ADMM}, the operator $\mathcal{P}_{\mathrm{psd}}$ denotes the projection onto the positive semidefinite cone $\bs^+$, and is computed via  eigendecomposition (see \cite{Wen2010} for details).
ADMM typically converges in a reasonable time to a solution with acceptable accuracy.
The number of ADMM iterations required for convergence depends on the desired accuracy as well as the formation graph (e.g., when the formation graph is complete, it is straightforward to show that ADMM converges to the optimal solution in a single iteration).
The time comparisons between an existing SDP solver for \eqref{eq:OptimCVX} and the presented ADMM method \eqref{eq:ADMM} are given in Section~\ref{sec:exper}. 

\subsection{Robustness, Collision Avoidance, and Formation Size}\label{sec:robustness}

Gain matrices are recomputed only when a new desired formation is specified.
During execution, vehicles use the gains and the relative position of their neighbors to compute the velocity vector $u_i$ in \eqref{eq:control} at each time instance. Having $u_i$ computed, the vehicle's low-level controller is tasked with tracking the direction and speed specified by this vector.

One can show that 1) \textit{any} \textit{positive} scaling; and 2) \textit{any} rotation less than \textit{90 degrees} of the velocity vector $u_i$ does not void the convergence guarantees of the formation control strategy (see \cite[Thm. 2]{Fathian2019}). 
These key properties indicate extreme robustness to errors and disturbances. For example, discrepancies between the actual and desired velocity of a vehicle caused by imperfect tracking, unmodeled dynamics, or small misalignments in $z$-axes of UAV body frames can be modeled as a positive scaling and small rotation of the nominal $u_i$, for which  convergence to the desired formation is unaffected.
The aforementioned properties can be further used for collision avoidance, where velocity vectors that lead vehicles to close proximity are modified to prevent collisions. More specifically, \eqref{eq:control} can be modified as
\begin{gather} \label{eq:RotationCtrl}
u_i \eqdef c_i \, R_i \sum_{j \in \mathcal{N}_i}  A_{ij} \, (q_j - q_i),
\end{gather}
where the rotation matrix $R_i$, which is limited to $90$ degrees, is chosen to rotate any velocity vector that brings two vehicles closer than a specified distance. 
If there is no feasible direction of motion within this range, the scalar $c_i$, which is normally set to one, is set to zero to stop the vehicle.

The collision avoidance strategy \eqref{eq:RotationCtrl} runs onboard, but comes at the cost of losing convergence guarantees since vehicles can become gridlocked due to the unavailability of motion directions (allowing $c_i = 0$ in \eqref{eq:RotationCtrl} violates the aforementioned property in which $c_i > 0$ is required to ensure convergence). 
Optimal assignment of vehicles to target formation points guarantees non-intersecting lines from their current positions to their assigned points (see \cite[Thm. 3.1]{Turpin2014}).
\edit{Since the vehicle body frames can be nonaligned and the control is distributed, vehicles are not expected to move perfectly in a straight line under our control strategy. However, assignment can still help deconflict the swarm by increasing the availability of motion directions.
The impact of assignment on resolving gridlocks is shown in Section~\ref{sec:exper}. }

Finally, note that \eqref{eq:RotationCtrl} leads to achieving the desired formation \textit{shape}, but the formation \textit{size} is not regulated and depends on the initial position of the vehicles. To control the size, \eqref{eq:RotationCtrl} can be augmented to contract (or expand) the formation when the vehicles are farther (or closer) than the desired distance. 
This augmented strategy, and its theoretical convergence guarantees, is discussed in our earlier work for 2D formations (see (74) in \cite{Fathian2018b}). Since the extension to 3D formations considered in this work is straightforward, we omit this discussion for brevity.

\section{\edit{Distributed} Task Assignment} \label{sec:assign}

The goal of task assignment is to uniquely allocate each UAV to a point in the desired formation.
A natural objective for this task is to minimize the overall distance from the UAVs to their assignments in the desired formation.
We are interested in the final 3D geometric shape rather than the exact location and yaw of the end formation.
To this effect, we allow a rotation $R$ around the $z$-axis, and translation $t$ of the desired formation coordinates that minimize the overall distance from the UAVs to the rotated and translated desired formation.
More precisely, our objective is to find the assignment map $\sigma$ that solves
\begin{equation}\label{eq:assign_prob}
\begin{aligned}
\underset{\substack{R\in\mathcal{R}_z,\;t\in\br^3\\\sigma\in\mathrm{S}_n}}{\text{minimize}} && \sum_{i=1}^{n} \| q_i - (R \, p_{\sigma(i)}+t) \|^2,
\end{aligned}
\end{equation}
where $q_i$ denotes the UAV positions, 
$\mathrm{S}_n$ is the symmetric group of all permutations from the set $\{1,\dots, n\}$ to itself,
and $\mathcal{R}_z$ is the set of rotation matrices around the $z$-axis. Recall that the 
$z$-axes of UAV body frames are assumed to be aligned, as per Section~\ref{sec:motionplanning}.

\edit{Three challenges arise in finding a distributed solution for~\eqref{eq:assign_prob}.
First, the objective of \eqref{eq:assign_prob} includes the positions of all UAVs, whereas each UAV only obtains the positions of its neighbors.
Second, the UAV body frames are not aligned and the UAVs only know the transformations between their body frames and their neighbors.  
Lastly, \eqref{eq:assign_prob} is a non-convex mixed integer program, for which finding the global optimizer becomes intractable for large $n$.} 
As computational efficiency and scalability are of utmost concern for UAV platforms, 
we settle with obtaining a suboptimal answer via a coordinate descent approach and an auction assignment strategy.
This approach is inspired by \cite{macdonald2011multi}, which in contrast uses a centralized Hungarian algorithm for assignment.
Every iteration of our algorithm consists of an \textit{alignment} stage and an \textit{assignment} stage, where the assignment is fixed as we solve for an alignment, and vice versa.

\subsection{Alignment}

\begin{figure} [t!]
	\centering
	\includegraphics[width=\columnwidth]{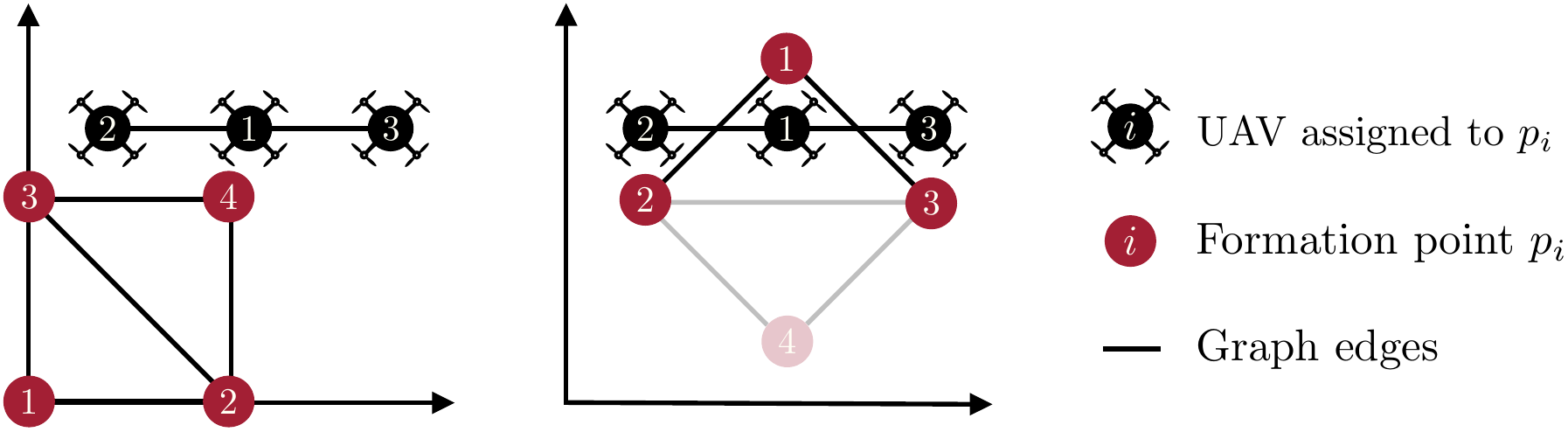}    
		\caption{
		Illustrative 2D alignment example with four vehicles from UAV~1's perspective.
        UAV $4$ is not shown because it does not communicate with UAV $1$.
		\textbf{(left)}~New formation graph, UAV $1$ and its neighbors before the alignment.
		\textbf{(right)}~Aligned formation 
		based on UAV $1$, its neighbors and their corresponding formation points. 
		The formation point associated to UAV 4 is faded to indicate that UAV 1 does not have information about it.
		}	
	\label{fig:alignment}
\end{figure}

Given an assignment $\sigma^*$ (e.g., identity assignment ${\sigma^*(i)=i}$ for every new formation, or prior assignment computed for the same formation), UAV $i$ solves a \edit{distributed} formulation of \eqref{eq:assign_prob} given by
\begin{equation}\label{eq:assign_prob_fix_sigma}
\begin{aligned}
\underset{\substack{R_i\in \mathcal{R}_z,\;t_i\in\br^3}}{\text{minimize}} && \sum_{j\in\mathcal{N}'_i} \| q_j - (R_i \, p_{\sigma^*(j)} + t_i) \|^2,
\end{aligned}
\end{equation}
where ${\mathcal{N}}'_i \eqdef \mathcal{N}_i\cup \{i\}$, and positions $q_j$ are in UAV $i$'s start frame.
In \eqref{eq:assign_prob_fix_sigma}, UAV $i$ aims to align the desired formation to minimize the distance to its own and neighbors' positions based on the given assignment $\sigma^*$.   
Fig.~\ref{fig:alignment} gives an illustrative example of this stage.
Problem \eqref{eq:assign_prob_fix_sigma} is the well-known point cloud alignment problem, for which the optimal solution $\left(R^*_i,t^*_i\right)$ 
is obtained from Arun's method~\cite{arun1987least}  using the projection of $q_j$ and $p_{\sigma^*(j)}$ on the {$x$-$y$ plane} for the rotation. %

\subsection{Assignment}

In this stage, the UAVs aim to collaboratively find an assignment based on the results obtained from their alignment stage. The assignment problem is formulated as
\begin{equation}\label{eq:assign_prob_fix_Rt}
\begin{aligned}
\underset{\substack{\sigma\in\mathrm{S}_n}}{\text{minimize}} && \sum_{i=1}^{n} \| q_i - (R_i^* \, p_{\sigma(i)}+t_i^*) \|^2.
\end{aligned}
\end{equation}
Problem~\eqref{eq:assign_prob_fix_Rt} is a linear sum assignment problem that can be solved optimally by methods such as the distributed Hungarian algorithm~\cite{giordani2010distributed, chopra2017distributed}, which converges in $\text{O}(n^3)$ iterations. 
Due to onboard resource constraints, we trade optimality for computational efficiency by using our prior work, the consensus-based auction algorithm (CBAA)~\cite{choi2009consensus}. 
CBAA is guaranteed to converge in at most $n d$ iterations, where $d$ is the diameter of the formation graph, $\mathcal{G}$.

To bring \eqref{eq:assign_prob_fix_Rt} into the standard form for applying CBAA, let binary variables $x_{ij}$ represent the assignment $\sigma$, where ${x_{ij} = 1}$ if ${\sigma(i) = j}$ and $0$ otherwise.
Further, let ${c_{ij} \eqdef 1\,/\,\| q_i-(R_i^* \, p_j + t_i^*) \|^2}$ denote the positive score for assigning UAV $i$ to formation point $j$. In practice, a small positive number can be added to the denominator of $c_{ij}$ to avoid division by zero.
It is straightforward to show that~\eqref{eq:assign_prob_fix_Rt} can be expressed as the integer program
\begin{equation}\label{eq:cbaa_problem}
\begin{aligned}
\underset{x_{ij} \in \{0,1\}}{\text{maximize}} &
&&\sum_{i,j=1}^{n} c_{ij} \,x_{ij}, \\
\text{subject to}
&&& \textstyle\sum_{i=1}^{n} x_{ij} = 1, \; \forall_{j} \\
&&& \textstyle\sum_{j=1}^{n} x_{ij} = 1,  \; \forall_{i} 
\end{aligned}
\end{equation}
where the constraints on $x_{ij}$ enforce conflict-free and one-to-one assignment captured by $\sigma \in S_n$ in~\eqref{eq:assign_prob_fix_Rt}, and maximizing~\eqref{eq:cbaa_problem} is equivalent to minimizing the overall distance from the UAVs to the rotated and translated formation in~\eqref{eq:assign_prob_fix_Rt}.

In executing CBAA, UAV $i$ stores and updates its own assignment and a list of winning bids (initialized as zeros) for all formation points. Each iteration of CBAA consists of an auction phase and a consensus phase. 
In the auction phase, UAV $i$ determines which formation point it would like to be assigned to in three steps: (1) check if any formation point $p_j$ produces a score $c_{ij}$ higher than its current winning bid; (2) of those formation points, set $x_{ij}=1$ for the $p_j$ that produces the highest score; (3) update the bid for the winning $p_{j}$ with new score $c_{ij}$.
In the consensus phase, vehicles converge on a common winning bid list. UAV $i$ exchanges its winning bid list with its neighbors and updates its list with the highest values from its own and all received lists. It sets $x_{ij}=0$ if the new winning bid for $p_j$ is higher than $c_{ij}$, implying that a different vehicle has been assigned to $p_j$.

Since CBAA is distributed, no central authority exists to affirm convergence.
Therefore, we enforce a synchronous execution to terminate the algorithm in $n d$ iterations, which is the maximum number of iterations required to guarantee convergence. 
The final assignment is recovered by letting $\sigma^*(i)=j$ for each $x_{ij}=1$.
CBAA guarantees a conflict-free assignment even though UAVs \edittwo{do not have a common reference frame} and may have inconsistent position estimates or different $R^*_i$ and $t^*_i$ for alignment.
\edittwo{We emphasize that although $c_{ij}$ is calculated by each UAV using only local knowledge, CBAA assigns UAVs to formation points without conflict.}
Further, it retains at least $50\%$ of the optimal performance; that is, given the optimal overall score $C^*$ of \eqref{eq:cbaa_problem} and the $C$ resulted from CBAA, $C/C^* \geq 0.5$.

\begin{table}[t!]	
    \ra{1.1}
	\centering
	\caption{Execution time of the CVX solver used for \eqref{eq:OptimCVX} vs. our ADMM solver \eqref{eq:ADMM} for obtaining formation gains for different number of vehicles. Reported times are in seconds and rounded to two decimals.}
	\begin{tabularx}{\columnwidth}{@{} l rrrrr @{}} %
		\toprule
		Algorithm  &  \multicolumn{5}{c @{}}{Number of Vehicles}\\ 
		\cmidrule(l){2-6}
		& {5} & {20} & {50} & {100} &  {200}  \\
		\midrule
		CVX-SDP time         & 0.54 & 32.48 & 8684.24 &  \texttt{OOM} & \texttt{OOM}  \\
		ADMM time (ours) & 0.01 & 0.03 & 1.31 & 12.26 & 134.67  \\
		\bottomrule
	\end{tabularx}
	\\ [0.2em]
	\begin{flushleft}
	{\scriptsize \texttt{OOM}: Out of memory}
	\end{flushleft}
	\label{tbl:gaintimes}
	\vskip-0.05in
\end{table}

\section{Experimental Results} \label{sec:exper}

This section shows that our distributed formation control and \edit{distributed} task assignment solutions scale with the number of UAVs, resolve gridlocks resulting from collision avoidance, and reduce the total distance traveled.

First, we investigate scalability by comparing the runtime of our ADMM-based solver~\eqref{eq:ADMM} with the interior-point method used in CVX (\href{http://cvxr.com/cvx}{\texttt{http://cvxr.com/cvx}}) to solve the SDP formulation~\eqref{eq:OptimCVX}.
These results are shown in Table~\ref{tbl:gaintimes}, and are generated in MATLAB using an Intel Core i7-7700K with \SI{32}{\giga\byte} RAM.
While the interior-point method becomes intractable for formations with more than 50 vehicles, our ADMM approach can solve for the control gains in seconds.

Second, we use software-in-the-loop simulations and hardware demonstrations to highlight how task assignment leads to quicker formation convergence with nearly \SI{100}{\percent} success.
Our pipeline is implemented in \texttt{C++} using Robot Operating System (ROS)~\cite{Quigley2009}.
Hardware demonstrations use a team of custom-built hexarotors, each with a diameter of \SI{0.5}{\meter} and an all-up-weight of \SI{1.1}{\kilo\gram}.
Code runs onboard the Qualcomm Snapdragon Flight board that includes a platform-optimized VIO package that outputs odometry at \SI{30}{\hertz} \cite{VIO}.
For simplicity of the implementation and the safety of the vehicles, we use our localization module (see Fig.~\ref{fig:System}) to inform each vehicle of every other vehicle's position.
However, the information about non-neighbors is only used for collision avoidance and could alternatively be found using, for example, onboard cameras.

\begin{table}[t]
\centering
\ra{1.1}
\caption{
Simulation results for 30 vehicles over 100 Monte Carlo trials.
Using our distributed assignment algorithm, we obtain results closer to the optimal, but centralized, Hungarian approach.
}
\label{tbl:simresults}
\begin{tabular}{l@{\hskip3pt}r x{1cm}x{0.6cm}x{1cm}x{0.6cm} r}
\toprule
&& \multicolumn{2}{c}{Distance Traveled (m)} & \multicolumn{2}{c}{Convergence Time (s)} & Success \\ \cmidrule{3-6}
&& mean & std & mean & std & \\ \midrule
\multirow{2}{*}{\normalsize{NA}}
& nc &
28.2 & 4.1 & 131.0 & 30.0 & \SI{58}{\percent} \\
& c    &
28.6 & 3.6 & 134.0 & 25.2 & \SI{66}{\percent} \\ [0.2em]
\multirow{2}{*}{\normalsize{A}}
& nc &
10.9 & 2.2 & 64.7 & 38.6 & \SI{98}{\percent} \\
& c    &
9.9 & 2.0 & 68.1 & 43.7 & \SI{96}{\percent} \\ [0.2em]
\multirow{1}{*}{\normalsize{H}}
& c    &
5.1 & 1.0 & 40.6 & 53.5 & \SI{100}{\percent} \\
\bottomrule
\end{tabular}
\\ [0.2em]
\begin{flushleft}
{\scriptsize NA: no assignment\quad A: \edit{distributed} assignment (ours)\quad H: centralized Hungarian} \\
{\scriptsize c: complete graph\quad nc: non-complete graph}
\end{flushleft}
\vskip-0.05in
\end{table}

\subsection{Simulations}

We perform Monte Carlo trials to measure the impact of \edit{distributed} task assignment on a large team of vehicles.
A trial consists of randomly initializing 30 UAVs in a \SI[product-units = single]{20 x 20}{\meter} area, where the minimum distance between initial positions is \SI{1.5}{\meter}.
A random formation is generated for each trial within a \SI[product-units = single]{15 x 15 x 2}{\meter} volume, with a minimum distance between formation points of \SI{2}{\meter}.
A trial is completed once the swarm has successfully reached the formation from the random initialization.
If the swarm is trapped in a gridlock for more than \SI{90}{\second}, the trial is considered indefinitely gridlocked and is aborted.

For each trial, our pipeline is tested in three main configurations: with centralized assignment, with distributed assignment, and without assignment.
Except for centralized assignment, each configuration is further tested with both a complete and randomly generated non-complete formation graph.
Centralized assignment provides an optimal baseline for comparison and is performed using the Hungarian algorithm with a complete graph.
The assignment algorithms are executed at a period of \SI{2}{\second}, allowing the swarm to resolve gridlocks by enabling new collision-free motion directions.

\begin{figure}[t]
    \centering
    \includegraphics[width=1\columnwidth]{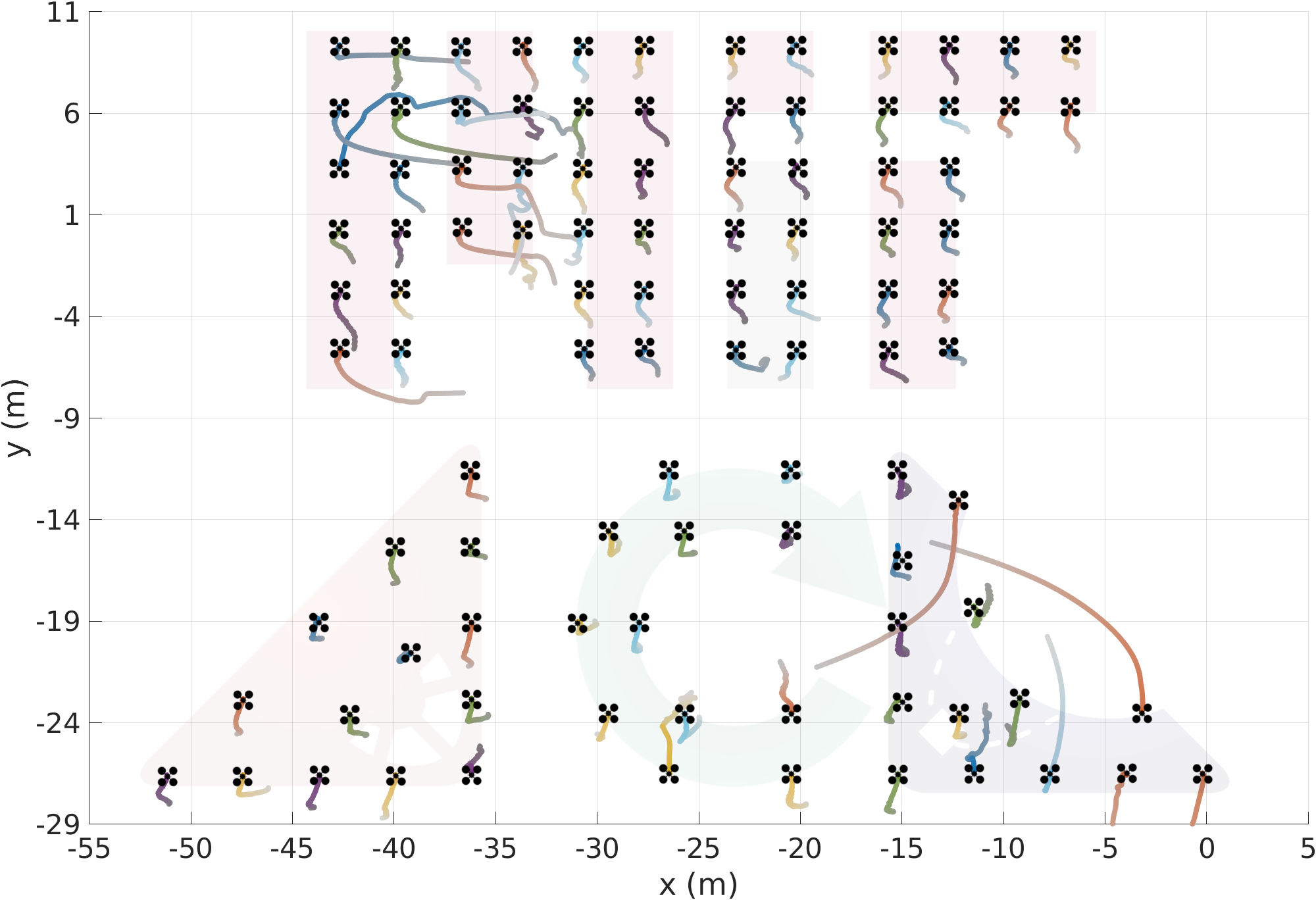}
    \caption{Large-scale simulation with 100 UAVs. The last \SI{40}{\second} of motion are shown and UAVs are depicted at 2x scale for better visibility.}
	\label{fig:mitacl100}
\end{figure}

Table~\ref{tbl:simresults} shows the comparison results, where for successful trials the average distance traveled and average flying time to converge to the desired formation are reported.
As expected, the centralized Hungarian approach obtains \SI{100}{\percent} success rate with the shortest distance traveled \edit{and only an average of 2.0 reassignments} to converge to the formation.
\edit{However, this approach has a computational cost of $O(n^3)$ in the number of vehicles and relies on a centralized coordinator in a common reference frame with complete knowledge of the swarm (Fig.~\ref{fig:frame-a}).
On the other hand, our CBAA-based assignment algorithm is a more scalable deconfliction strategy that is executed in the non-aligned frames of the UAVs (Fig.~\ref{fig:frame-c}) and is nearly optimal in practice as confirmed by the \SI{97}{\percent} average convergence rate in Table~\ref{tbl:simresults}, with an average of 11.6 reassignments}.
Compared to formation control without assignment, our algorithm allows the swarm to achieve formation convergence in nearly every case and in half the amount of time, on average.
The results also show that, on average, there is no significant performance decrease between complete and non-complete formation graphs.
Thus, non-complete formation graphs can be used to reduce communication overhead without sacrificing the convergence rate or ability to reach the desired formation.

\edit{We remark that symmetric formations may lead our task assignment strategy~\eqref{eq:assign_prob} to exhibit momentary swapping behavior.
However, noise in each UAV's sensing is included in the simulation and we have not observed convergence failure of~\eqref{eq:assign_prob} in simulation or hardware.
We believe this swapping behavior is caused by ignoring the vehicle dynamics in~\eqref{eq:assign_prob} and consider this in future work.
}

To demonstrate scalability, we performed a large-scale simulation with 100 UAVs randomly initialized in a \SI[product-units = single]{60 x 30}{\meter} area.
This simulation was performed using Amazon Web Services.
Vehicles achieve the \texttt{MIT ACL} formation using a sparse formation graph with only \SI{24}{\percent} of the edges in a complete graph, which is beneficial for bandwidth-limited communication.
The last \SI{40}{\second} are shown in Fig.~\ref{fig:mitacl100}, where the motion traces indicate deconfliction due to reassignment.

\begin{figure}[t!]
	\centering
	\begin{subfigure}[b]{0.32\columnwidth}
	    \centering
	    \includegraphics[width=\textwidth, trim={3cm 0.5cm 3cm 2cm},clip]{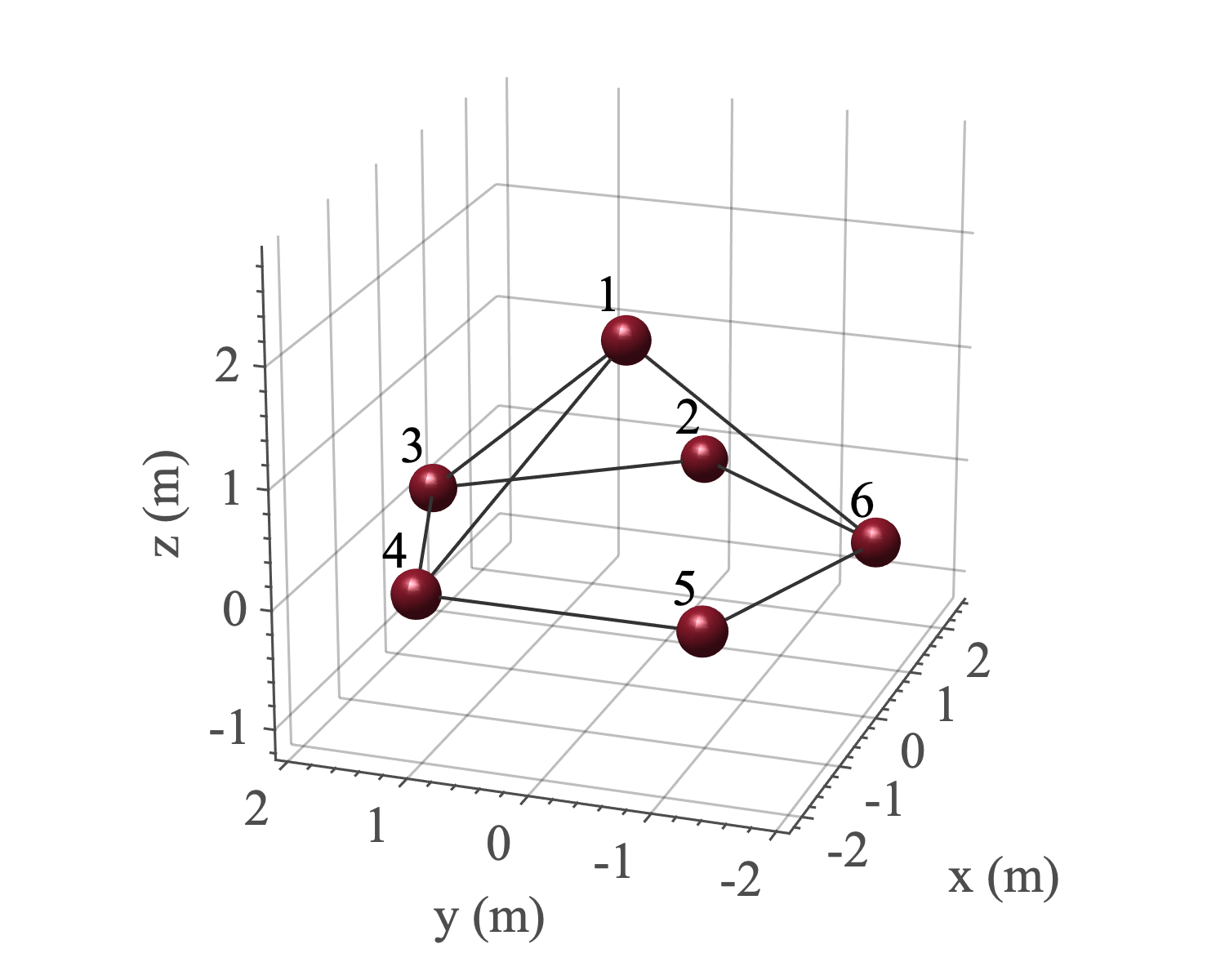}
	    \caption{\scriptsize Pentagonal pyramid}
	\end{subfigure}
	\begin{subfigure}[b]{0.32\columnwidth}
	    \centering
	    \includegraphics[width=\textwidth, trim={3cm 0.5cm 3cm 2cm},clip]{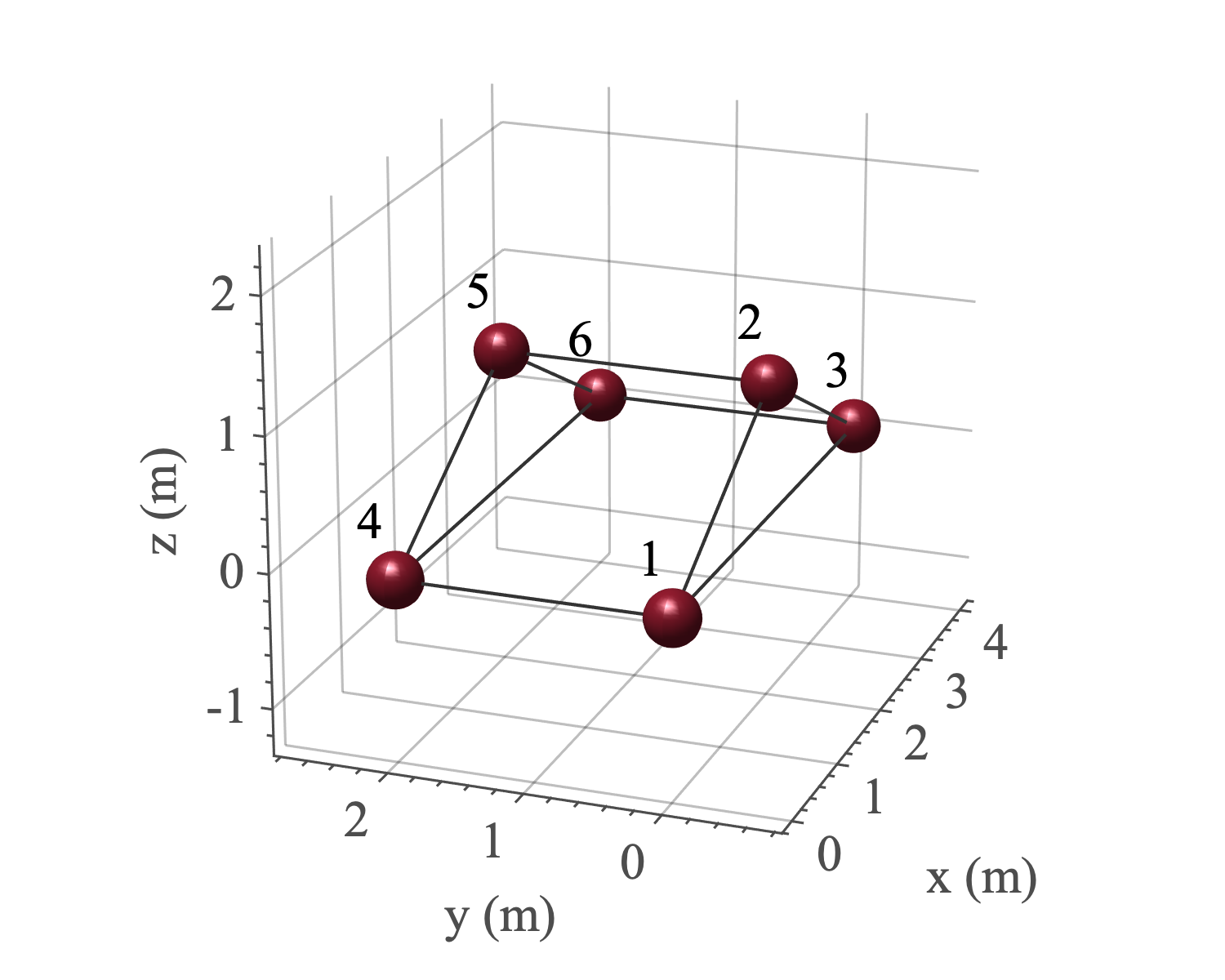} 
	    \caption{\scriptsize Triangular prism}
	\end{subfigure}
	\begin{subfigure}[b]{0.32\columnwidth}
	    \centering
	    \includegraphics[width=\textwidth, trim={3cm 0.5cm 3cm 2cm},clip]{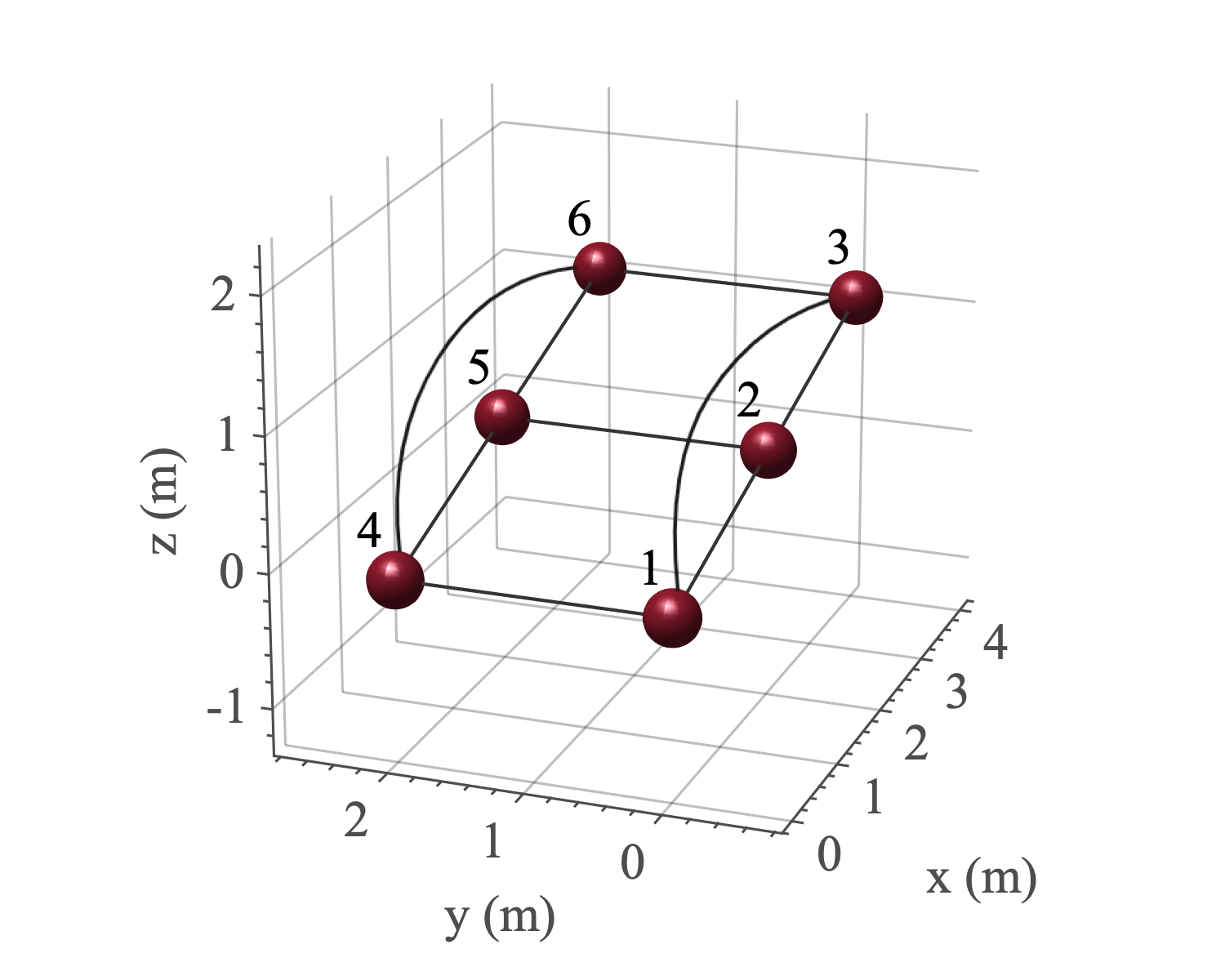} 
	    \caption{\scriptsize Slanted plane}
	\end{subfigure}
	\caption{Non-complete formation graphs used in the hardware experiments.}
	\label{fig:exp_formations}
\end{figure}

\begin{figure}[t]
    \centering
    \includegraphics[trim=20 85 20 120, clip, width=1\columnwidth]{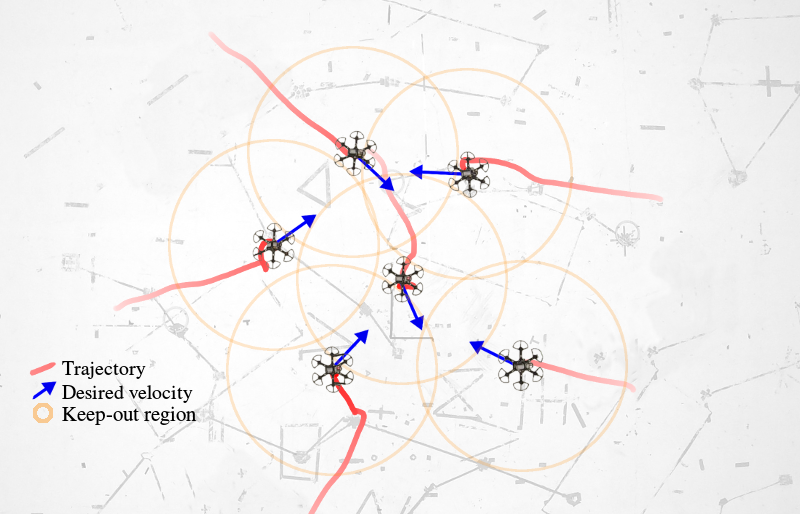}
    \caption{Without assignment, UAVs attempting to achieve the pyramid formation are gridlocked due to collision avoidance.}
	\label{fig:hwgridlock}
\end{figure}

\subsection{Hardware Demonstrations}

We demonstrate formation flight with six UAVs by cycling through the three formations illustrated in Fig.~\ref{fig:exp_formations}.
The minimum distance between desired formation points is \SI{2}{\meter} for each formation.
\edit{Because the time required to calculate the formation gains from~\eqref{eq:ADMM} is small, in our experiments each UAV independently calculates the formation gains onboard in \SI{20}{\milli\second}.
In this case, the base station is used only to dispatch the desired formation graph to the UAVs.}

The UAVs are initialized at pre-specified locations so that the transforms between vehicles' VIO start frames are known.
After taking off and hovering, an operator dispatches each desired formation to the swarm.
Four configurations are tested by cycling through the formations twice.
For each configuration, a total of six trials are recorded and averaged over, where a trial is the transition from the current swarm state to the next desired formation.
When assignment is used, the period of reassignment is \SI{1.2}{\second}.

Consistent with the simulations, the results in Table~\ref{tbl:hwresults} indicate that without our assignment strategy, the vehicles fail to achieve the desired formation in up to \SI{50}{\percent} of the trials, while every trial using assignment was successful.
An example of convergence failure is shown in Fig.~\ref{fig:hwgridlock}.

The supplementary video provides insights into the qualitative behavior of our system.
Note that the achieved formations in the video are occasionally inverted from the desired formations shown in Fig.~\ref{fig:exp_formations}.
Recall that our formation control aims to achieve the desired \textit{shape}.
The inverted formations seen in experiments are due to negative scaling in the $z$-axis.
We also point out that the formations shown in Fig.~\ref{fig:exp_formations} are not universally rigid. 
Universal rigidity is a sufficient condition for our gain design, but not necessary.
In practice, formations with sparser graphs can be used,
so long as the recovered gain matrix leads to a negative objective~\eqref{eq:OptimX}.
This helps to alleviate communication load across the swarm.

\edit{The transmission requirements for localization and assignment are $5.2$ kbps per neighbor and $0.064nd(n+1)$ kb per neighbor at the reassignment period, respectively.
For example, in our experiments with non-complete graphs, the theoretical bandwidth between each vehicle is approximately $9$ kbps.
Using a sparse graph, a reassignment period of \SI{30}{\second}, and mid-grade WiFi connectivity, the expected upper bound before channel saturation is 800 UAVs.
These numbers are supported by our simulation of 30 UAVs in a non-complete formation graph, where we measured 2161 kbps of communication between a UAV and its neighbors.
}

\begin{table}[t]
\centering
\ra{1.1}
\caption{
Hardware results.
Our distributed assignment algorithm successfully breaks gridlock and converges to every desired formation.
}
\label{tbl:hwresults}
\begin{tabular}{l@{\hskip3pt}r x{1cm}x{0.6cm}x{1cm}x{0.6cm} r}
\toprule
&& \multicolumn{2}{c}{Distance Traveled (m)} & \multicolumn{2}{c}{Convergence Time (s)} & Success \\ \cmidrule{3-7}
&& mean & std & mean & std & \\ \midrule
\multirow{2}{*}{\normalsize{NA}}
& nc &
0.8 & 0.5 & 15.5 & 9.2 & \SI{50}{\percent} \\
& c    &
0.9 & 1.0 & 11.3 & 3.0 & \SI{67}{\percent} \\ [0.2em]
\multirow{2}{*}{\normalsize{A}}
& nc &
1.4 & 0.8 & 14.3 & 8.7 & \SI{100}{\percent} \\
& c    &
0.8 & 0.6 & 10.1 & 4.8 & \SI{100}{\percent} \\
\bottomrule
\end{tabular}
\\ [0.2em]
\begin{flushleft}
{\scriptsize NA: no assignment\quad A: \edit{distributed} assignment (ours)} \\
{\scriptsize c: complete graph\quad nc: non-complete graph}
\end{flushleft}
\vskip-0.1in
\end{table}

\section{Conclusion and Future Work}

We presented a unified formation flying pipeline with distributed formation control and task assignment solutions that run onboard the vehicles and uses VIO for localization. 
Our ADMM solver addressed the scalability issue of general solvers for obtaining formation gains and the auction-based algorithm generated non-conflicting assignment solutions in a computationally efficient manner. 
Simulation and hardware tests demonstrated formation convergence in \SIrange[range-units=single,range-phrase=--]{96}{100}{\percent} of cases that gridlocked when assignment was not used.
Noteworthy future extensions include incorporating an assignment strategy that considers vehicle dynamics to minimize the total \textit{predicted} distance traveled, and addition of distributed pose graph optimization to obtain consistent VIO pose estimates.

\balance %

\bibliographystyle{IEEEtran}
\bibliography{Bibs}

\begin{thebibliography}{10}
\providecommand{\url}[1]{#1}
\csname url@samestyle\endcsname
\providecommand{\newblock}{\relax}
\providecommand{\bibinfo}[2]{#2}
\providecommand{\BIBentrySTDinterwordspacing}{\spaceskip=0pt\relax}
\providecommand{\BIBentryALTinterwordstretchfactor}{4}
\providecommand{\BIBentryALTinterwordspacing}{\spaceskip=\fontdimen2\font plus
\BIBentryALTinterwordstretchfactor\fontdimen3\font minus
  \fontdimen4\font\relax}
\providecommand{\BIBforeignlanguage}[2]{{%
\expandafter\ifx\csname l@#1\endcsname\relax
\typeout{** WARNING: IEEEtran.bst: No hyphenation pattern has been}%
\typeout{** loaded for the language `#1'. Using the pattern for}%
\typeout{** the default language instead.}%
\else
\language=\csname l@#1\endcsname
\fi
#2}}
\providecommand{\BIBdecl}{\relax}
\BIBdecl

\bibitem{VIO}
\url{https://developer.qualcomm.com/software/machine-vision-sdk}.

\bibitem{Wang2017}
L.~Wang, A.~Ames, and M.~Egerstedt, ``Safety barrier certificates for
  collisions-free multirobot systems,'' \emph{IEEE TRO}, vol.~33, no.~3, pp.
  661--674, 2017.

\bibitem{Fathian2019}
K.~Fathian, S.~Safaoui, T.~H. Summers, and N.~R. Gans, ``Robust {3D}
  distributed formation control with collision avoidance and application to
  multirotor aerial vehicles,'' \emph{IEEE ICRA}, pp. 9209--9215, 2019.

\bibitem{Preiss2017}
J.~Preiss, W.~Honig, G.~Sukhatme, and N.~Ayanian, ``Crazyswarm: A large
  nano-quadcopter swarm,'' in \emph{IEEE ICRA}, 2017, pp. 3299--3304.

\bibitem{Honig2018}
W.~H{\"o}nig, J.~A. Preiss, T.~S. Kumar, G.~S. Sukhatme, and N.~Ayanian,
  ``Trajectory planning for quadrotor swarms,'' \emph{IEEE TRO}, vol.~34,
  no.~4, pp. 856--869, 2018.

\bibitem{Du2019}
X.~Du, C.~Luis, M.~Vukosavljev, and A.~Schoellig, ``Fast and in sync: Periodic
  swarm patterns for quadrotors,'' in \emph{IEEE ICRA}, 2019, pp. 9143--9149.

\bibitem{wilson2020robotarium}
S.~Wilson, P.~Glotfelter, L.~Wang, S.~Mayya, G.~Notomista, M.~Mote, and
  M.~Egerstedt, ``{The Robotarium: Globally Impactful Opportunities,
  Challenges, and Lessons Learned in Remote-Access, Distributed Control of
  Multirobot Systems},'' \emph{IEEE CSM}, vol. 40(1), pp. 26--44, 2020.

\bibitem{enright2004spheres}
J.~Enright, M.~Hilstad, A.~Saenz-Otero, and D.~Miller, ``{The SPHERES guest
  scientist program: Collaborative science on the ISS},'' in \emph{IEEE
  Aerospace Conference Proceedings}, vol.~1, 2004.

\bibitem{Forster2013}
C.~Forster, S.~Lynen, L.~Kneip, and D.~Scaramuzza, ``{Collaborative monocular
  SLAM with multiple micro aerial vehicles},'' in \emph{IEEE/RSJ IROS}, 2013,
  pp. 3962--3970.

\bibitem{Loianno2016}
G.~Loianno, Y.~Mulgaonkar, C.~Brunner, D.~Ahuja, A.~Ramanandan, M.~Chari,
  S.~Diaz, and V.~Kumar, ``A swarm of flying smartphones,'' in \emph{IEEE/RSJ
  IROS}, 2016, pp. 1681--1688.

\bibitem{Weinstein2018}
A.~Weinstein, A.~Cho, G.~Loianno, and V.~Kumar, ``Visual inertial odometry
  swarm: An autonomous swarm of vision-based quadrotors,'' \emph{IEEE RA-L},
  vol.~3, no.~3, pp. 1801--1807, July 2018.

\bibitem{Oh2015}
K.-K. Oh, M.-C. Park, and H.-S. Ahn, ``A survey of multi-agent formation
  control,'' \emph{Automatica}, vol.~53, pp. 424--440, 2015.

\bibitem{Delmerico2018}
J.~Delmerico and D.~Scaramuzza, ``A benchmark comparison of monocular
  visual-inertial odometry algorithms for flying robots,'' in \emph{IEEE ICRA},
  2018, pp. 2502--2509.

\bibitem{Turpin2014}
M.~Turpin, N.~Michael, and V.~Kumar, ``Capt: Concurrent assignment and planning
  of trajectories for multiple robots,'' \emph{IJRR}, vol.~33, no.~1, pp.
  98--112, 2014.

\bibitem{van2011reciprocal}
J.~Van Den~Berg, S.~J. Guy, M.~Lin, and D.~Manocha, ``Reciprocal n-body
  collision avoidance,'' in \emph{Robotics research}.\hskip 1em plus 0.5em
  minus 0.4em\relax Springer, 2011, pp. 3--19.

\bibitem{morgan2016swarm}
D.~Morgan, G.~P. Subramanian, S.-J. Chung, and F.~Y. Hadaegh, ``Swarm
  assignment and trajectory optimization using variable-swarm, distributed
  auction assignment and sequential convex programming,'' \emph{The
  International Journal of Robotics Research}, vol.~35, no.~10, pp. 1261--1285,
  2016.

\bibitem{Montijano2016}
E.~Montijano, E.~Cristofalo, D.~Zhou, M.~Schwager, and C.~Saguees,
  ``Vision-based distributed formation control without an external positioning
  system,'' \emph{IEEE TRO}, vol.~32, no.~2, pp. 339--351, 2016.

\bibitem{Tron2016}
R.~Tron, J.~Thomas, G.~Loianno, K.~Daniilidis, and V.~Kumar, ``A distributed
  optimization framework for localization and formation control: Applications
  to vision-based measurements,'' \emph{IEEE CSM}, vol.~36, no.~4, pp. 22--44,
  Aug 2016.

\bibitem{arun1987least}
K.~S. Arun, T.~S. Huang, and S.~D. Blostein, ``Least-squares fitting of two
  {3-D} point sets,'' \emph{IEEE TPAMI}, no.~5, pp. 698--700, 1987.

\bibitem{Gortler2014}
S.~J. Gortler and D.~P. Thurston, ``Characterizing the universal rigidity of
  generic frameworks,'' \emph{Disc. \& Comp. Geom.}, vol.~51, no.~4, pp.
  1017--1036, 2014.

\bibitem{Lin2016}
Z.~Lin, L.~Wang, Z.~Han, and v.~Minyue~Fu, ``{A Graph Laplacian Approach to
  Coordinate-Free Formation Stabilization for Directed Networks},'' \emph{IEEE
  TAC}, vol.~61, no.~5, pp. 1269--1280, May 2016.

\bibitem{Lin2016a}
Z.~Lin, L.~Wang, Z.~Chen, M.~Fu, and Z.~Han, ``Necessary and sufficient
  graphical conditions for affine formation control,'' \emph{IEEE TAC},
  vol.~61, no.~10, pp. 2877--2891, 2016.

\bibitem{Fathian2018b}
K.~Fathian, S.~Safaoui, T.~H. Summers, and N.~R. Gans, ``Robust distributed
  planar formation control for higher-order holonomic and nonholonomic
  agents,'' \emph{arXiv preprint, arXiv:1807.11058}, 2018.

\bibitem{Ren2007}
W.~Ren, ``Consensus strategies for cooperative control of vehicle formations,''
  \emph{IET CTA}, vol.~1, no.~2, pp. 505--512, 2007.

\bibitem{Zhao2019}
S.~Zhao and D.~Zelazo, ``Bearing rigidity theory and its applications for
  control and estimation of network systems: Life beyond distance rigidity,''
  \emph{IEEE CSM}, vol.~39, no.~2, pp. 66--83, 2019.

\bibitem{Montijano2014}
E.~Montijano, D.~Zhou, M.~Schwager, and C.~Sagues, ``Distributed formation
  control without a global reference frame,'' in \emph{IEEE ACC}, 2014, pp.
  3862--3867.

\bibitem{lusk2020distributed}
P.~C. Lusk, X.~Cai, S.~Wadhwania, A.~Paris, K.~Fathian, and J.~P. How, ``A
  distributed pipeline for scalable, deconflicted formation flying,'' 2020,
  {\url{https://arxiv.org/abs/2003.01851}}.

\bibitem{Wen2010}
Z.~Wen, D.~Goldfarb, and W.~Yin, ``Alternating direction augmented lagrangian
  methods for semidefinite programming,'' \emph{Mathematical Programming
  Computation}, vol.~2, no. 3-4, pp. 203--230, 2010.

\bibitem{macdonald2011multi}
E.~A. Macdonald, ``Multi-robot assignment and formation control,'' Master's
  thesis, Georgia Institute of Technology, 2011.

\bibitem{giordani2010distributed}
S.~Giordani, M.~Lujak, and F.~Martinelli, ``A distributed algorithm for the
  multi-robot task allocation problem,'' in \emph{IEA/AIE}.\hskip 1em plus
  0.5em minus 0.4em\relax Springer, 2010, pp. 721--730.

\bibitem{chopra2017distributed}
S.~Chopra, G.~Notarstefano, M.~Rice, and M.~Egerstedt, ``A distributed version
  of the hungarian method for multirobot assignment,'' \emph{IEEE TRO},
  vol.~33, no.~4, pp. 932--947, 2017.

\bibitem{choi2009consensus}
H.-L. Choi, L.~Brunet, and J.~P. How, ``Consensus-based decentralized auctions
  for robust task allocation,'' \emph{IEEE TRO}, vol.~25, no.~4, pp. 912--926,
  2009.

\bibitem{Quigley2009}
M.~Quigley, K.~Conley, B.~Gerkey, J.~Faust, T.~Foote, J.~Leibs, R.~Wheeler, and
  A.~Y. Ng, ``{ROS: an open-source Robot Operating System},'' in \emph{ICRA
  workshop on open source software}, vol.~3, no. 3.2.\hskip 1em plus 0.5em
  minus 0.4em\relax Kobe, Japan, 2009, p.~5.

\end{thebibliography}

\appendix \label{sec:Appendix}

\begin{proof}[\textbf{Proof of Proposition~\ref{prop:SDP}}]
Consider problem \eqref{eq:OptimZ}. 
The facts that $B\, M = 0$ and $R \in \br^{n\times (n-2)}$ is the orthogonal complement of $M$ imply that $B$ can be factored as ${B = - R \, Z \, R^\top}$, where $Z \in \bs^+_{n-2}$. Substituting $B$ with ${- R \, Z \, R^\top}$ in \eqref{eq:OptimZ} and simplifying yields 
\begin{equation} \label{eq:OptimSDP1}
\begin{aligned}
& \underset{Z \in \bs^+_{n-2}}{\text{maximize}}
& & \lambda_{\text{min}}(Z) & \\
& \text{subject to}
& & \left[R \, Z \, R^\top \right]_{ij} = 0 \;   & \forall_{i} \ \forall_{j \notin \mathcal{N}_i} \\
&&& \tr(Z) = \text{constant} & \\
\end{aligned}
\end{equation}
which reduces the dimension of the optimization variable from $n$ to $n-2$.
Note that the constraint $B\, M = 0$ in \eqref{eq:OptimZ} is automatically satisfied in \eqref{eq:OptimSDP1} as $R^\top M = 0$ by orthogonality.
The objective of \eqref{eq:OptimSDP1}, i.e., maximizing the smallest eigenvalue of the positive semidefinite matrix $Z$, can be expressed equivalently as finding $Z$ and the smallest $\gamma \geq 0$ such that $Z - \, \gamma^{-1} I$ remains positive semidefinite (this statement can be proved by diagonalizing $Z$). Hence, \eqref{eq:OptimSDP1} can be expressed as
\begin{equation} \label{eq:OptimSDP2}
\begin{aligned}
& \underset{Z \in \bs^+_{n-2}}{\text{minimize}} 
& & \gamma  \\
& \text{subject to}
& & \gamma \geq 0, \quad
Z - \, \gamma^{-1} I \succeq 0  \\
&&& \left[R \, Z \, R^\top \right]_{ij} = 0 \;   & \forall_{i} \ \forall_{j \notin \mathcal{N}_i} \\
&&& \tr(Z) = \text{constant} & \\
\end{aligned}
\end{equation}
Let $C \eqdef \begin{bsmallmatrix}
I & 0 \\
0 & 0
\end{bsmallmatrix}$ 
and 
$X \eqdef \begin{bsmallmatrix}
\gamma\, I && I \\
I && Z 
\end{bsmallmatrix}$,
where the size of the identity matrix $I$ is the same as $Z$. The Schur complement condition for positive semidefinite matrices states that ${X \succeq 0}$ if and only if $\gamma\, I \succeq 0$ and $Z - I \, (\gamma \, I)^{-1} I \succeq 0$. The latter implies that $\gamma \geq 0$ and $Z - \gamma^{-1} I \succeq 0$, which are the constraints in \eqref{eq:OptimSDP2}. Consequently, \eqref{eq:OptimSDP2} can be written concisely as 
\begin{equation} \label{eq:OptimSDP3}
\begin{aligned}
& \underset{X \in \bs^+_{2n-4} }{\text{minimize}} 
& & \langle C,\, X \rangle \\
& \text{subject to}
& & \sa(X) = b
\end{aligned}
\end{equation}
where $\langle C,\, X \rangle$ is the Frobenius inner product, and $\sa(X) = b$ captures the linear constraints on both the structure of $X$, i.e., the identity blocks and the last two constraints in~\eqref{eq:OptimSDP2}.
\end{proof}

\end{document}